\documentclass{article}
\usepackage{spconf,amsmath,graphicx}

\usepackage{amssymb}
\usepackage{dsfont}
\usepackage{amsfonts}
\usepackage{mathrsfs}

\usepackage{setspace}
\usepackage{xspace}

\usepackage{ stmaryrd }
\usepackage{lineno}
\usepackage{color}
\usepackage{url}
\usepackage{ stmaryrd }
\usepackage{kmath}

\usepackage{amssymb,color} 

\def\x{{\mathbf x}}

\title{Reduced-Order Modeling Of Hidden Dynamics}
\name{Patrick H\'eas and C\'edric Herzet}
\address{INRIA Centre Rennes - Bretagne Atlantique, Campus universitaire de Beaulieu, 35000 Rennes, France}
\newcommand{\Rr}{\mathds{R}}

\newcommand{\dd}{{u}}

\newcommand{\D}{\mathbf{u}}

\newcommand{\y}{\mathbf{y}}

\newcommand{\eg}{\textit{e.g.}, }
\newcommand{\ie}{\textit{i.e.}, }

\DeclareMathOperator*{\argmin}{arg\,min}

\usepackage{soul} 

\begin{document}
%
\maketitle
\begin{abstract}

The objective of this paper is to investigate how  noisy and incomplete observations can be integrated in the process of building a reduced-order model. 
 This problematic arises in many scientific domains where there exists a  need  for accurate low-order descriptions of highly-complex phenomena, which can not be directly and/or deterministically observed. 
Within this context, the paper proposes a probabilistic framework for the construction of \textit{``POD-Galerkin''} reduced-order models. Assuming a hidden Markov chain, the inference integrates  the uncertainty of the hidden states relying on their posterior distribution.  
Simulations show the benefits obtained by exploiting the proposed framework.   
\end{abstract}

\section{Introduction} \vspace{-0.1cm}

In many fields of Sciences, one is interested in studying the {spatio-temporal} evolution of a state variable characterized by  a  differential equation.  Numerical discretization in space and time leads to a high dimensional system of equations {of the form:} \vspace{-0.35cm}
\begin{align}\label{eq:model_init} 
 \left\{\begin{aligned}
& x_{t+1}= f_t(x_t,\theta_t) , \\
&x_1=\theta_1,
\end{aligned}\right. \vspace{-0.25cm}
\end{align} 
\noindent
 {where}  $x_t\in\Rr^n$ {is the spatial discretization of the state variable at time $t$,}  $f_t:\Rr^n \times \Rr^{p_t} \to  \Rr^n$  and    $\theta_t \in \Rr^{p_t}$  {denotes some parameters}. 
As a few examples of systems obeying this type of constraints, one can mention the wave equation characterizing the propagation of sound  \cite{Godlewski96}, the Navier-Stokes equations describing fluid evolution \cite{quartapelle13} or the Maxwell's equations governing the realm of electromagnetism \cite{Godlewski96}.
 {Because \eqref{eq:model_init} may correspond to a very high-dimensional system in some applications, computing a trajectory $\{x_t\}_t$ given  some parameters $\{\theta_t\}_t$}
  may  lead to unacceptable computational burdens.




To deal with this computational bottleneck, the concept of ``reduced-order models'' (ROMs) has been introduced in many communities dealing with high-dimensional   systems. The idea of ROMs is fairly simple: one wishes to find a  system involving a small number of degrees of freedom (typically ten to one hundred) while allowing for a reasonable characterization of the  state of the   high-dimensional  system, in a certain range of {operating} regimes. 
Due to their paramount practical importance, the construction of ROMs has a fairly long history in the community of experimental physics and geophysics, beginning with the pioneer{ing} works of Lorenz  in the 60's \cite{Lorenz56,Lorenz60,Lorenz63}. Among others, one can mention  Hankel norm approximations \cite{green12},   principal orthogonal decomposition combined with Galerkin projections  (POD-Galerkin) \cite{Homescu07},  principal oscillating patterns or principal interacting patterns \cite{Hasselmann88}. We refer the reader to the book by Antoulas for a review~\cite{Antoulas05}.

The set of operating regimes to be reproduced by a ROM   is of the form \vspace{-0.15cm}
\begin{align*}
\mathcal{X}\triangleq \{\x \triangleq (x_1 \cdots x_T) :  \x \mbox{ obeys \eqref{eq:model_init} with $\{\theta_t\}_{t=1}^T\in \mathcal{R}$ }  \},
\end{align*}
where  $\mathcal{R}\subseteq \Rr^{p_1} \times \cdots \times \Rr^{p_T}$.
 In words,  
 the set of regimes $\cal{X}$ is defined by the set $\mathcal{R}$ of admissible parameters, which determines  state trajectories through recursion  \eqref{eq:model_init}.  The set $\mathcal{X}$ may either be  known perfectly or only partially.  In the first situation,   the construction of a ROM can rely on the perfect characterization of a representative set of trajectories,  as soon as recursion \eqref{eq:model_init} can be computed for the  known regimes, see \eg \cite{Sirovich87}.  Building a ROM in the second situation is more subtle:   representative trajectories can not be computed directly by \eqref{eq:model_init}  since there exists an uncertainty on the regimes {of interest}. For example, in fluid mechanics, the initial and the boundary conditions, defining the evolution of the fluid, are rarely perfectly known.
A third  situation, which is even  more dramatic, emerges when recursion \eqref{eq:model_init} is intractable due to  a prohibitive computational burden. This  typically happens in climate studies, where it is  inconceivable to use high-dimensional numerical simulations for determining global warming scenarios.

A useful ingredient in  these uncertain and/or intractable contexts   is to incorporate {partial} observations {of} {$x_t$'s}  in the process of ROM construction. For example,  in geophysics, satellites   daily provide a huge  quantity of observations on the  ocean or the atmosphere evolution. In the literature,  a common strategy {consists in}  substituting the representative trajectories (intractable to compute in these cases) by a set of observations representative of the regimes  \cite{Moreno04,Dadamo07,vonStorch01}. 
Nevertheless, this straightforward approach is  in most experiments flawed. It ignores that observations may be incomplete and affected by noise which may  dramatically  impact the  ROM.

In this paper, we propose a new model-reduction technique which: \textit{(i)} accounts for the uncertainties in the system to reduce; \textit{(ii)} exploits  observations  in the reduction process  while taking into account   their imperfect nature.
 We focus on the family of ROMs based on POD-Galerkin projection {and recast the latter within a probabilistic framework}.
 {In a nutshell, our approach is based on a probabilistic characterization of the uncertainties of the high-dimensional system given the data, and the exploitation of this posterior information  in the reduction process.}
We illustrate the proposed approach by the reduction of a system of 2D Navier-Stokes equations.  \vspace{-0.cm}

\section{ A Posteriori  Inference Problem}\vspace{-0.1cm}
 
\subsection{Surrogate Prior and Observation Models} \vspace{-0.1cm}
 Our first step will consist in defining a high-dimensional probabilistic model gathering the uncertainty on the state of the system. 
 Because we are assuming that there are some uncertainties in the set of admissible parameter $\mathcal{R}$ (\ie the set of operating regimes $\mathcal{X}$), we will suppose that each $\theta_t$ in \eqref{eq:model_init} is the realization of {a} random variable $\Theta_t$ distributed according to a probability measure $\kappa_t$ of support $\mathcal{R}$.   This  implies that $x_t$ is seen as a realization of a random variable $X_t$ distributed according to some probability measure $\nu_t$ of support $\mathcal{X}$.
 
As mentioned previously,  the problem is that we are unaware of   $\mathcal{X}$, since we do not have a perfect knowledge of   $\mathcal{R}$. 
 {Assume now we know a measure $\eta_t$ dominating $\nu_t$, that is for any element $A$ of the Borel sets  in $\Rr^{n}$, $\eta_t(A)=0$ implies $\nu_t(A)=0$.}  {Since $\nu_t$ is unknown, we will use $\eta_t$ as a \textit{surrogate} measure on the unknown operating regimes.}  {Because of the domination relation, state trajectories belonging to the unknown set $\mathcal{X}$ will have a non-zero probability. } 
 We  further assume that the $\eta_t$'s {are characterized by} a probabilistic model of the form:\vspace{-0.15cm}
   \begin{align}
& \left\{\begin{aligned} \label{eq:recOrginalNon}
  & X_t=b_t( X_{t-1})+V_t, \quad  V_ t\sim \zeta^v_t(dv_t),\\
&X_1 \sim \eta_1(dx_1), 
\end{aligned}\right.
\end{align}
where we have introduced  some  operator  $b_t: \Rr^n \to  \Rr^n$ and where the $V_t$'s are mutually independent {random variables}  of realization $v_t\in \Rr^n$ and of probability measure $\zeta^v_t$.
%
The definition of {\eqref{eq:recOrginalNon}}  usually stems from 
the inclusion of some  knowledge on the physics  and  the nature of the uncertainties. Indeed, there often exist  approximated probabilistic characterizations of deterministic chaotic systems, see \eg \cite{RobertVargas08,Chevillard10,Heas14,Heas13} for turbulent systems.   

On top of {a prior information}  {$\eta_t$} on the ${X}_t$, we assume that we have at our disposal a set of $M$ observations on the states of the system{, say $\{Y^i_t\}_{i=1}^M$}.
  For a sequence of $T$ state variables observed $M$ times, we define the random matrices  $\mathbf{{X}} \triangleq (X_1 \cdots X_T)$ and $\mathbf{{Y}} \triangleq (Y^1_1 \cdots Y^1_T\,Y^2_1 \cdots Y^M_T)$ of realizations $\x=(x_1 \cdots x_T)  ~\in ~\Rr^{n\times T}$ and $\mathbf{y}=(y^1_1 \cdots y^1_T\, y^2_1 \cdots y^M_T)  \in \Rr^{m\times T M}$.    Assume   $X_t$'s and $Y^i_t$'s satisfy \vspace{-0.1cm}
  \begin{align}
Y^i_t = h_t( X_t)+W_t, \quad     W_t \sim \zeta^w_t(dw_t),\label{eq:recYOrginalNon}
\end{align}
where we have introduced  some   operator  ${h}_t : \Rr^{n}\to \Rr^{m}$ and where the $W_t$'s are mutually independent noises  of realization $w_t\in \Rr^m$ and of probability measure $\zeta^w_t$. 
Given this set of observations, one can hope to 
 remove certain uncertainties on the system regime, the {final} goal being to include this information in the model reduction process. 
 In the following,  we will be interested in Bayesian estimators relying on the joint posterior measure of  $\mathbf{X}$ given some observation $\mathbf{Y}=\mathbf{y}$, say $\mu$. The posterior measure will be associated to the hidden Markov model (HMM) defined by \eqref{eq:recOrginalNon} -  \eqref{eq:recYOrginalNon}.  For HMMs, the posterior admits the factorization~\cite{DelMoral12} \vspace{-0.15cm}
  \begin{align}\label{eq:post}
\mu(d\x,\mathbf{y})=\frac{g(\x,\mathbf{y})\eta(d\x)}{\langle g(\cdot,\mathbf{y}),\eta\rangle },
\end{align} 
with the prior
$\eta(d\x)\triangleq \eta_1(dx_1)\prod_{t=2}^T \zeta^v_t(dx_t-b_t(x_{t-1})),$ 
and the likelihood $ g(\x,d\mathbf{y})\triangleq\prod_{t=1,i=1}^{T,M} \zeta^w_t(dy^i_t-h_t(x_t))$, where  the expectation  for any integrable function $\varphi$ is  \vspace{-0.15cm}
\begin{align*}
 \langle \eta,\varphi \rangle \triangleq \int_{\Rr^{n \times T} } \eta (d\x) \varphi(d\x).
\end{align*}
Let us remark that, under specific conditions, a Bayesian estimator is asymptotically efficient \cite{Lehmann03}. In other words,  for $M$ sufficiently large, the effect of the prior probability  \eqref{eq:recOrginalNon} on the posterior is negligible.   \vspace{-0.2cm}


\subsection{Uncertainty-Aware POD-Galerkin  Projection} \vspace{-0.1cm}
The low-rank approximation called \textit{Galerkin} projection  of the dynamics \eqref{eq:model_init} 
is  obtained by   projecting  $x_t$'s onto a subspace spanned by the columns of some matrix $\D\in \Rr^{n\times k}$ where $k < n$ \cite{Homescu07}. More precisely, it consists in a recursion \vspace{-0.15cm}
\begin{align}\label{eq:redModel}
 \left\{\begin{aligned}
&z_t=\D^*{f}_t(\D z_{t-1}, \theta_{t-1}),\\
&z_1=\D^*\theta_{1}, 
\end{aligned}\right.\vspace{-0.25cm}
\end{align} 
 \noindent implying  a sequence of $k$-dimensional variables  $\{z_t \in\Rr^k\}_t$, where the exponent $^*$ denotes the conjugate transpose. {Because $k<n$}, system \eqref{eq:redModel} is usually tractable. {Once recursion \eqref{eq:redModel} has been evaluated, an approximation of state $x_t$ can be obtained by a matrix-vector multiplication $\D z_t$}.

There  exist several criteria {in the literature} to choose matrix $\D$ \cite{Antoulas05}.   The \textit{POD-Galerkin} approximation infers matrix $\D$ by minimiz{ing} the cost  $\phi:\Rr^{n \times T} \times  \Rr^{n \times k} \to \Rr_+$  \vspace{-0.1cm}
\begin{align}\label{eq:costError}
\phi(\x ,\D)\triangleq \parallel \x  -\D\D^*\x  \parallel^2_F,
\end{align}
 over the set of unitary matrices 
 $\mathcal{U}=\{\D \in \Rr^{n\times k}| \D^*\D=\mathbf{i}_k\},$
 where $\mathbf{i}_k$ denotes the $k$-dimensional identity matrix. 
 
In the context of our probabilistic modeling, $\x$ is a realization of some random variable governed by  \eqref{eq:recOrginalNon} which is  indirectly observed through  \eqref{eq:recYOrginalNon}. The uncertainty on the state $\x$ given some observation $\mathbf{Y}=\mathbf{y}$ is  quantified by the posterior \eqref{eq:post}.   
Using this information, we define the inference of matrix $\D$ for POD-Galerkin projection as the solution of
\begin{align}\label{eq:romCrit3Last}
\argmin_{\D \in \mathcal{U}} \langle  \mu(\cdot,\mathbf{y}),\phi(\cdot,\D) \rangle.  \vspace{-0.15cm}
\end{align}  
{We discuss the resolution of this problem in the next section.}\vspace{-0.25cm}

\section{Computational Considerations}\vspace{-0.15cm}

\subsection{Solution in Terms of Posterior Expectation} \vspace{-0.1cm}

While being non-convex, problem  \eqref{eq:romCrit3Last}  admits a closed-form solution in terms of posterior expectation. Let $\{\sigma_j\}_{j=1}^{n}$ and  $\{\hat \dd_j\}_{j=1}^{n}$  {respectively denote} the eigenvalues  and eigenvectors of   \vspace{-0.35cm}
\begin{align}\label{eq:2ndMoment}
\langle \mu(d\x,\mathbf{y}), \x\x^*\rangle,
\end{align}
 with $\sigma_j\ge \sigma_{j+1}$.  For any $k$-dimensional unitary basis $\D$, it {is} straightforward to deduce that the expected cost in \eqref{eq:romCrit3Last} admits the  lower bound 
$\sum_{j=k+1}^{n} \sigma_j \le   \langle  \mu(\cdot,\mathbf{y}),\phi(\cdot, {\D}) \rangle, 
$ 
and that this bound is reached for the matrix $\hat \D$ whose columns are 
  the eigenvectors $\{\hat \dd_j\}_{j=1}^{k}$   associated to the  $k$ largest eigenvalues  $\{\sigma_j\}_{j=1}^{k}$, see \cite{eckart36, Horn12}.
Expectation  \eqref{eq:2ndMoment} can be decomposed as  \vspace{-0.45cm}
\begin{align}\label{eq:approxSecMoment}
\langle \mu(d\x,\mathbf{y}),  \x\x^* \rangle=
 \sum_{t=1}^{T} \mathbf{ p}_t+\bar{x}_t(\y) \bar{x}_t^*(\y). 
\end{align}
where  $\bar x_{t}(\y) \in \Rr^{n}$ and $\mathbf{p}_{t}  \in \Rr^{n\times n}$ {are the a posteriori  mean  and covariance of $x_t$ given $\y$}. In the particular case of Gaussian linear  or finite state HMMs,  
the posterior mean and covariance are explicitly   given by  \textit{Kalman} recursions  or by the \textit{Baum-Welsh} re-estimation  formulae \cite{Briers10}.  In the general case,   expectations with respect to the posterior measure \eqref{eq:post} do not admit closed-form expressions. Nevertheless, sequential Monte-Carlo methods  provide asymptotically consistent estimators \cite{DelMoral12}.

We highlight the relevance  of the proposed approach by a comparison with the standard  \textit{snapshot method} \cite{Sirovich87}. This method  consists in  
{computing $ \mathbf{\hat u}$ as the $k$ first eigenvectors of }\vspace{-0.2cm}
\begin{align}\label{eq:approxSecMomentSnapshot}
 \sum_{t=1}^{T} \hat{x}_t(\y) \hat{x}_t^*(\y),
\end{align}
where $ \hat{x}_t(\y)$ denotes some {estimate} of the state  $x_t$ given observations  $\mathbf{y}$.
  Consequently, the snapshot method  ignores directions where the covariance of the estimation error is large due to incomplete or noisy observations. In particular, for the minimum mean square error estimator  $\hat{x}_t(\y)= \bar{x}_t(\y) $, it neglects the influence of $\mathbf{p}_t$ in   \eqref{eq:approxSecMoment}. With the proposed approach,  the columns of $\hat \D$ will  also be chosen in the directions in which  {the eigenvalues of} $\mathbf{p}_t$ are large, \ie directions  where {some} uncertainty remains {a posteriori}. \vspace{-0.25cm}

\subsection{Complexity Issues and Krylov Approximations}\vspace{-0.15cm}

Assuming we can compute   matrix \eqref{eq:2ndMoment}, which is usually full rank, the  complexity necessary to solve exactly the eigenvalue problem scales   as $\mathcal{O}(n^2k)$  \cite{Golub96}. \vspace{-0.05cm}
Since the goal is to lower the problem dimension, one is inevitably faced to a prohibitive dimensionality $n$. We  choose to resort to 
Krylov {subspace} approximations \cite{Antoulas05}. 
 In a nutshell, an approximation in Krylov subspaces enables to represent a large matrix   by some approximated singular value decomposition, where  the $k$ eigenvectors associated to the largest  eigenvalues are {approximated} relying on  the Arnoldi iteration method~\cite{Arnoldi51}.
 The method complexity scales as $\mathcal{O}(k^2n)$, \ie present{s} the great advantage to be linear with respect to the state variable dimension. It is comparable to the  complexity $\mathcal{O}(T(T^2+n))$  required  by the diagonalisation of matrix \eqref{eq:approxSecMomentSnapshot} of rank $T$ and by the  matrix-vector multiplications in the  snapshot method. \vspace{-0.cm}



\section{Numerical Example}\label{sec:4}\vspace{-0.15cm}

\subsection{Reduction of 2D Navier-Stokes Equations}

\subsubsection{2D Turbulence Model}\label{sec:turbMod} \vspace{-0.1cm}
We illustrate the proposed methodology within the context of reducing a 2D model of turbulence from a video depicting the evolution of a scalar field transported by the flow. 
As described in  \cite{Carlier05}, the  high-dimensional model  \eqref{eq:model_init} is here  a quadratic function  with respect to {a spatial discretization of the flow velocity} $x_t\in \Rr^n$ \vspace{-0.15cm}
\begin{align}\label{eq:turb2D}
 f_t(x_t,\theta_t)= \mathbf{c}^*(\mathbf{i}_n+\alpha\boldsymbol{\ell}-x_t^*\mathbf{r})\mathbf{c}x_t +\theta_t , 
 \end{align}
where  $\alpha\in \Rr^+$ denotes a dissipation coefficient  and where matrices $\boldsymbol{\ell}\in \Rr^{m\times m}$, $\mathbf{c}\in \Rr^{m\times n}$ and $\mathbf{r}\in \Rr^{n \times m}$  with $n=2m$ are discrete versions of respectively the Laplacian, the curl  and the spatial gradient operators. Variable $\theta_t \in \Rr^{n}$ accounts for some  forcing. The observation model relates the flow {velocity} $x_t$ {to the variation $y_t\in \Rr^m$ of the intensity  of a scalar field  conveyed by the flow.}  It  takes the form of \eqref{eq:recYOrginalNon} with a zero-mean uncorrelated Gaussian noise $W_t$ of variance $\sigma^2$ and with a linear function    \vspace{-0.15cm}
\begin{align}
h_t(x_t)=\mathbf{h}_tx_{t}+\xi_t,
 \end{align}
  for some matrix $\mathbf{h}_t \in \Rr^{m\times n}$ and vector $\xi_t\in \Rr^m$, see    \cite{Carlier05}.
{We use} the high-dimensional model {to} produce one sequence of {2D  motion  field $\{x_t\}_{t=1}^T\in \mathcal{X}$  for a given operating regime $\{\theta_t\}_{t=1}^T \in \mathcal{R}$ and one  sequence ($M=1$) of scalar fields of intensity variations $y^1_t$'s}. {In the sequel, we denote these sequences as $(\x^{dns},\y^{dns}) \in \Rr^{n \times T} \times  \Rr^{m\times T}$.}   \vspace{-0.25cm}

\subsubsection{Optic-Flow Posterior} \vspace{-0.15cm}

As suggested  in the  literature of  \textit{optic-flow}  and turbulence modeling, we will rely on a degenerated case of  {the surrogate} prior \eqref{eq:recOrginalNon}   where $b_t(X_{t-1})$ vanishes. Moreover, we assume that $V_ t$'s  are zero-mean Gaussian random fields.  
Such priors are commonly used  in computer vision  \cite{Horn81},  in  fluid mechanics \cite{Heas14}, or  in geophysics \cite{Heas13} to describe  the correlated structure of motion fields.  These  priors  fulfill  the domination condition. They are usually degenerated and defined by an inverse covariance $\mathbf{q}_t\in \Rr^{n\times n}$ of rank lower than $n$.  The latter matrix implements typically some local regularity constraints, \eg finite difference  approximation of spatial gradients \cite{Horn81},  or some  self-similar  constraints and long-range dependency  proper to turbulent flows \cite{Heas14,Heas13}. In this linear Gaussian setting,
 we obtain  the  posterior mean and covariance 
\begin{equation}\label{eq:monoResMeanCov}
 \left\{\begin{aligned}
&\bar x_{t}(\y) =\sigma^{-2}\mathbf{  p}_{t}\mathbf{h}_t^*(y^1_{t}-\xi_t),\\
 &\mathbf{ p}_{t} =\sigma^{2}  (\mathbf{h}_t^*\mathbf{h}_t+\sigma^{2}\mathbf{q}_t)^{-1}.\vspace{-0.25cm}
  \end{aligned}\right.\vspace{-0.25cm}
\end{equation}
\noindent

\subsection{Experimental Setting, Results and Discussion} \vspace{-0.1cm}

%

Our simulations use a sample composed  of   $T=50$ consecutive states and images of size $m=2^7 \times 2^7$. 
We  evaluate  two different choices for $\mathbf{q}_t$: the standard gradient model \cite{Horn81} and the self-similar model proposed in  \cite{Heas13}. 
The inversion of the $n\times n$ matrices  appearing in \eqref{eq:monoResMeanCov} are approximated {within a linear complexity}  in a $10$-dimensional Krylov subspace.  
 The   $k=50$ first eigenvectors of \eqref{eq:approxSecMoment} composing the columns of   the minimizer $\hat \D$ of  \eqref{eq:romCrit3Last} are then approximated using a Krylov subspace of dimension $100$.

\begin{figure}	
\begin{tabular}{cc}
\includegraphics[width=.225\textwidth]{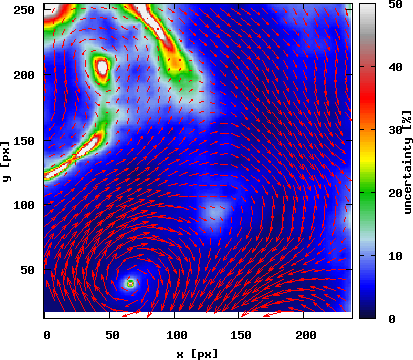}&\includegraphics[width=.225\textwidth]{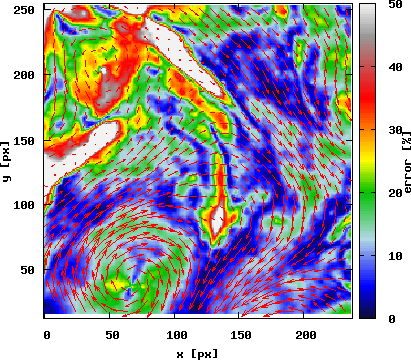}\\
\includegraphics[width=.225\textwidth]{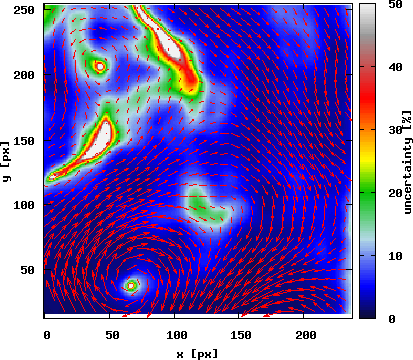}&\includegraphics[width=.225\textwidth]{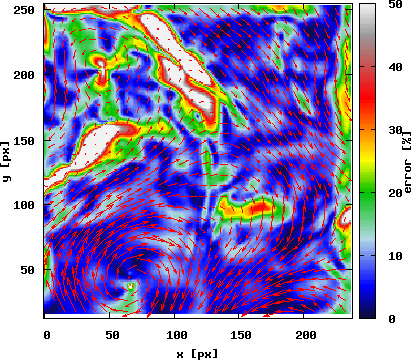}\vspace{-0.4cm}
\end{tabular}
	\caption{ \footnotesize{ \textbf{Left:} mean of the posterior bi-variate vector  {(arrows)} and   Frobenius norm of its  covariance   {(color map)} at each point of the pixel grid {for $t=25 $} . Results for standard  \cite{Horn81} (above) or self-similar prior \cite{Heas13} (below). \textbf{Right:}   ground truth bi-variate  vector {(arrows)} and   error \eqref{eq:errorMotionPixel}   {(color map)}  for the standard (above) or self-similar (below) prior. }
	 \label{fig:1}}\vspace{-0.15cm}
\end{figure}
\vspace{-0cm}
Fig. \ref{fig:1} illustrates the Gaussian posterior distribution of a turbulent flow obtained by optic-flow modeling, using Krylov approximation. 
The local posterior covariance (in fact the Frobenius norm of the local $2 \times 2$ covariance matrix)   can be compared to  the normalized squared  $\ell_2$-norm  error between the   realization $\x^{dns}\triangleq (x_1^{dns}\cdots x_T^{dns})$   and  the estimated posterior mean, \ie \vspace{-0.1cm}
\begin{align}\label{eq:errorMotionPixel}
{\| (x^{dns}_t)_s- (\bar x_t(\mathbf{y}^{dns}))_s \|^2_2}/{\| (x^{dns}_{t})_s \|^2_2},
\end{align}
where the subscript $s$ denotes the vector (bi-variate) component related  to the $s$-th pixel of the image grid.
  A visual inspection of the different maps shows that :
1) the region with high variances correspond to  areas characterized by large errors \eqref{eq:errorMotionPixel}; 
2) conversely, in the case of the self-similar prior, most regions with  large  errors \eqref{eq:errorMotionPixel} correspond  to areas with high variances; besides, the  error is  globally lower for the self-similar prior.
The first observation serves as {a} clear evidence of  the relevance of integrating the posterior distribution in the ROM building process. The second remark shows the importance of using good prior {surrogates}.  
\begin{figure}	
\begin{center}
\begin{tabular}{c}
\includegraphics[width=.35\textwidth]{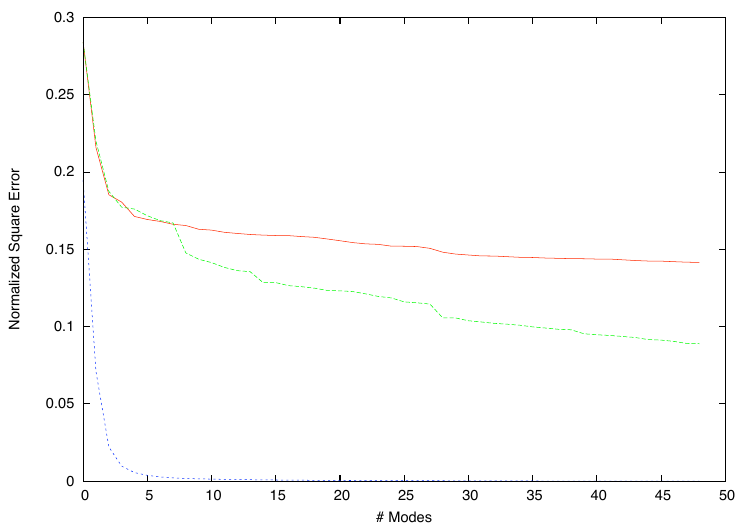}\vspace{-0.7cm}
\end{tabular}
\end{center}
	\caption{ \footnotesize{Evolution of the normalized  error \eqref{eq:errorRec} 
	 with respect to the dimension $k$ of the inferred reduced basis with the state-of-the-art snapshot method  \cite{Sirovich87}  (red solid  line), the proposed method (green dashed line) and using  directly the ground truth (blue dotted line).}
	 \label{fig:2}}\vspace{-0.15cm}
\end{figure}
\vspace{-0cm}

 The   plot of Fig. \ref{fig:2} shows  the  evolution of the normalized square  reconstruction error {induced}   by low-rank approximation of the true flow, \ie \vspace{-0.2cm}
\begin{align}\label{eq:errorRec}
{\| \x^{dns}- \hat\D\hat \D^*\x^{dns} \|^2_F}/{\| \x^{dns} \|^2_F},
\end{align}
with respect to the  number of columns $k$ of matrix $\hat \D$,  the  solution of \eqref{eq:romCrit3Last} using the prior model  in \cite{Horn81}. We evaluate the snapshot method and the proposed posterior reduced-order modeling. 
Moreover,  the figure also includes the plot of the error \eqref{eq:errorRec} obtained with a reduced basis $\hat \D$  inferred when $\x^{dns}$ is directly available without   uncertainty. We note that in principle, in order to evaluate the error induced by the ROM,  the orthogonal complement 
${\|  \hat\D\hat \D^*\x^{dns} -\hat\D\mathbf{z} \|^2_F}/{\| \x^{dns} \|^2_F}$ 
 should be added to error \eqref{eq:errorRec}, with $\mathbf{z}=(z_1\cdots z_T)$ computed from \eqref{eq:redModel} using $x_1^{dns}$  \cite{Homescu07}. Unfortunately, \cite{Carlier05} does  not provide the detailed implementation of  function \eqref{eq:turb2D}. 

We immediately remark that including uncertainty in the process of reduced-order modeling induces a clear gain in accuracy. The error reduction  is more than 30 $\%$.    
It is nevertheless modest in comparison to the gain  obtained by ROM  directly inferred  from ground truth. In the light of  comments of Fig. \ref{fig:1}, we  believe that this error gap may be significantly reduced by using more relevant priors, such as  those 
proposed in \cite{Heas14,Heas13}. {This is the topic of ongoing research.} \vspace{-0.cm}

\section{Conclusion} \vspace{-0.1cm}
 We have proposed a methodology for building ROMs which integrates the high-dimensional system uncertainties. It relies on the computation of an a posteriori measure. This information  reveals directions  where  uncertainty on the states remain high  and where  observations were insufficient to solve the regime ambiguities. We show how to integrate this a posteriori information {in the} construction of  ROMs {based on}  POD-Galerkin projections. 
Numerical experiments{, taking place in the context of 2D Navier-Stokes equations,} show that the proposed probabilistic {framework enables to}  lower significantly the reconstruction error in comparison to  {the standard ``POD-snapshot'' method.}  
  
%

 \newpage

\section*{Acknowledgements}
This work was
supported by the  ``Agence Nationale de la Recherche" through the GERONIMO project (ANR-13-JS03-0002).

\bibliographystyle{IEEEbib}
\bibliography{./bibtex}

\begin{thebibliography}{10}

\bibitem{Godlewski96}
E.~Godlewski and P.A. Raviart,
\newblock {\em Numerical Approximation of Hyperbolic Systems of Conservation
  Laws},
\newblock Number n¡~118 in Applied Mathematical Sciences. Springer, 1996.

\bibitem{quartapelle13}
L.~Quartapelle,
\newblock {\em Numerical Solution of the Incompressible Navier-Stokes
  Equations},
\newblock International Series of Numerical Mathematics. Birkh{\"a}user Basel,
  2013.

\bibitem{Lorenz56}
E.N. Lorenz,
\newblock ``Empirical orthogonal functions and statistical weather
  prediction,''
\newblock {\em Scientific Report 1, Statistical Forecasting Project, MIT,
  Cambridge, MA}, 1956.

\bibitem{Lorenz60}
E.N. Lorenz,
\newblock ``Maximum simplification of the dynamic equations,''
\newblock {\em Tellus}, vol. 12, no. 3, pp. 243--254, 1960.

\bibitem{Lorenz63}
E.~N. {Lorenz},
\newblock ``{Deterministic Nonperiodic Flow.},''
\newblock {\em Journal of Atmospheric Sciences}, vol. 20, pp. 130--148, Mar.
  1963.

\bibitem{green12}
M.~Green and D.J.N. Limebeer,
\newblock {\em Linear Robust Control},
\newblock Dover Books on Electrical Engineering. Dover Publications,
  Incorporated, 2012.

\bibitem{Homescu07}
C.~Homescu, L.~R. Petzold, and R.~Serban,
\newblock ``Error estimation for reduced-order models of dynamical systems,''
\newblock {\em {SIAM} Review}, vol. 49, no. 2, pp. 277--299, 2007.

\bibitem{Hasselmann88}
K.~Hasselmann,
\newblock ``Pips and pops: The reduction of complex dynamical systems using
  principal interaction and oscillation patterns,''
\newblock {\em Journal of Geophysical Research: Atmospheres}, vol. 93, no. D9,
  pp. 11015--11021, 1988.

\bibitem{Antoulas05}
A.C. Antoulas,
\newblock {\em Approximation of Large-scale Dynamical Systems},
\newblock Advances in Design and Control. Society for Industrial and Applied
  Mathematics, 2005.

\bibitem{Sirovich87}
L.~{Sirovich},
\newblock ``{Turbulence and the dynamics of coherent structures.},''
\newblock {\em Quarterly of Applied Mathematics}, vol. 45, pp. 561--571, Oct.
  1987.

\bibitem{Moreno04}
D.~Moreno, A.~Krothapalli, M.~B. Alkislar, and L.~M. Lourenco,
\newblock ``Low-dimensional model of a supersonic rectangular jet,''
\newblock {\em Phys Rev E (69: 026304)}, pp. 1--12, 2004.

\bibitem{Dadamo07}
J.~D'Adamo, N.~Papadakis, E.~Memin, and G.~Artana,
\newblock ``Variational assimilation of pod low-order dynamical systems,''
\newblock {\em Journal of Turbulence}, vol. 8, no. 9, pp. 1--22, 2007.

\bibitem{vonStorch01}
H.~von Storch and F.W. Zwiers,
\newblock {\em Statistical Analysis in Climate Research},
\newblock Cambridge University Press, 2001.

\bibitem{RobertVargas08}
R.~Robert and V.~Vargas,
\newblock ``Hydrodynamic turbulence and intermittent random fields,''
\newblock {\em Communications in Mathematical Physics}, vol. 284, pp. 649--673,
  2008.

\bibitem{Chevillard10}
L.~Chevillard, R.~Robert, and V.~Vargas,
\newblock ``A stochastic representation of the local structure of turbulence,''
\newblock {\em EPL (Europhysics Letters)}, vol. 89, no. 5, pp. 54002, 2010.

\bibitem{Heas14}
P.~Heas, F.~Lavancier, and S.~Kadri Harouna,
\newblock ``Self-similar prior and wavelet bases for hidden incompressible
  turbulent motion,''
\newblock {\em SIAM Journal on Imaging Sciences}, vol. 7, pp. 1171--1209, 2014.

\bibitem{Heas13}
P.~Heas, C~Herzet, E.~Memin, Heitz D., and P.D. Mininni,
\newblock ``Bayesian estimation of turbulent motion,''
\newblock {\em IEEE transactions on Pattern Analysis And Machine Intelligence},
  vol. 35, no. 6, pp. 1343--56, 2013.

\bibitem{DelMoral12}
P.D. Moral,
\newblock {\em Feynman-Kac Formulae: Genealogical and Interacting Particle
  Systems with Applications},
\newblock Probability and Its Applications. Springer New York, 2012.

\bibitem{Lehmann03}
E.L. Lehmann and G.~Casella,
\newblock {\em Theory of Point Estimation},
\newblock Springer Texts in Statistics. Springer New York, 2003.

\bibitem{eckart36}
R.M. Johnson,
\newblock ``On a theorem stated by \textit{Eckart} and \textit{Young},''
\newblock {\em Psychometrika}, vol. 28, no. 3, pp. 259--263, 1963.

\bibitem{Horn12}
R.~A. Horn and C.~R Johnson,
\newblock {\em Matrix analysis},
\newblock Cambridge university press, 2012.

\bibitem{Briers10}
M.~Briers, A.~Doucet, and S.~Maskell,
\newblock ``Smoothing algorithms for state--space models,''
\newblock {\em Annals of the Institute of Statistical Mathematics}, vol. 62,
  no. 1, pp. 61--89, 2010.

\bibitem{Golub96}
G.H. Golub and C.F. Van~Loan,
\newblock {\em Matrix Computations},
\newblock Johns Hopkins Studies in the Mathematical Sciences. Johns Hopkins
  University Press, 1996.

\bibitem{Arnoldi51}
W.~E. Arnoldi,
\newblock ``The principle of minimized iteration in the solution of the matrix
  eigenvalue problem,''
\newblock {\em Quarterly of Applied Mathematics}, vol. 9, pp. 17--29, 1951.

\bibitem{Carlier05}
J.~Carlier and B.~Wieneke,
\newblock ``Report 1 on production and diffusion of fluid mechanics images and
  data,''
\newblock {\em Fluid project deliverable 1.2.}, 2005.

\bibitem{Horn81}
B.~Horn and B.~Schunck,
\newblock ``Determining optical flow,''
\newblock {\em Artificial Intelligence}, vol. 17, pp. 185--203, 1981.

\end{thebibliography}

\end{document}